\documentclass[letterpaper, 10 pt, conference]{ieeeconf}  

\IEEEoverridecommandlockouts                              

\usepackage[T1]{fontenc}
\usepackage[latin9]{inputenc}
\usepackage{amstext}
\usepackage{graphicx}

\makeatletter
\@ifundefined{showcaptionsetup}{}{%
 \PassOptionsToPackage{caption=false}{subfig}}
\usepackage{subfig}
\usepackage{amsmath}
\usepackage{url}
\makeatother

\usepackage{babel}

\title{\LARGE \bf
Scalable Simulation and Demonstration of Jumping Piezoelectric 2-D Soft Robots
}

\author{Zhiwu Zheng, Prakhar Kumar, Yenan Chen, Hsin Cheng, \\
Sigurd Wagner, Minjie Chen, Naveen Verma and James C. Sturm
\thanks{This work was supported by the Semiconductor Research Corporation
(SRC), DARPA, Princeton Program in Plasma Science and Technology,
and Princeton University. \emph{(Corresponding author: Zhiwu Zheng)}}
\thanks{The authors are with the Department of Electrical and Computer Engineering,
Princeton University, Princeton, New Jersey 08544, U.S.A. (e-mails:
{zhiwuz@princeton.edu; prakhark@princeton.edu; yenanc@zju.edu.cn;
hsin@princeton.edu; wagner@princeton.edu; minjie@princeton.edu; nverma@princeton.edu;
sturm@princeton.edu}).}%
}

\begin{document}

\maketitle
\thispagestyle{empty}
\pagestyle{empty}

\begin{abstract}

Soft robots have drawn great interest due to their ability to take on a rich range
of shapes and motions, compared
to traditional rigid robots. However, the motions, and underlying statics and dynamics,
pose significant challenges to forming well-generalized and robust
models necessary for robot design and control. In this work, we demonstrate a five-actuator soft robot capable of complex motions and develop
a scalable simulation framework that reliably predicts robot motions. 
The simulation framework is validated
by comparing its predictions to experimental results, based on a robot constructed from piezoelectric layers bonded to a steel-foil substrate. The simulation framework exploits the physics engine PyBullet, and employs discrete rigid-link elements connected
by motors to model the actuators. We perform static and AC analyses
to validate a single-unit actuator cantilever setup and observe close agreement
between simulation and experiments for both the cases. The analyses
are extended to the five-actuator robot, where simulations accurately predict
the static and AC robot motions, including shapes for applied DC voltage inputs, nearly-static "inchworm" motion, and jumping (in vertical as well as vertical and horizontal directions). These motions exhibit complex non-linear behavior, with forward robot motion reaching \textasciitilde{}1 cm/s.
Our open-source code can be found at: \url{https://github.com/zhiwuz/sfers}.
\end{abstract}

\section{INTRODUCTION}

Soft robots have garnered interest because of their ability to take on
complex shapes and motions, especially involving rich interactions with their environments. There is growing interest
in leveraging the static and dynamic behavior of such robots, where for instance dynamics can enable significant speed enhancement by driving at mechanically resonant frequencies \cite{Wu2019,Ji2019}. This necessitates understanding the statics and dynamics through reliable models, as well as efficient integration and application of those models in simulators used
for robot design and development of control systems. However,
soft body modelling is challenging, due to the large number of degrees-of-freedom
and complicated interactions between soft bodies and the environment
(such as friction and collisions). Recent work on soft-robot modelling
primarily focuses on finite-element methods \cite{Duriez2013,Zhang2017,Goury2018,Zhang2019}
and/or pseudo-rigid body models \cite{Wu2019,Lobontiu2001,Bandopadhya2010,Best2016,Satheeshbabu2017,Li2018,DellaSantina2018,Tang2019}.

Most recent work focuses on pneumatic soft robots \cite{Duriez2013,Zhang2017,Goury2018,Best2016,Satheeshbabu2017,DellaSantina2018}
or shape-memory and motor-tendon actuators
\cite{Wang2014a,Umedachi2016}. Scalable approaches for electrostatic soft robots have been more limited, with some examples including pseudo-rigid-body based modelling of a single-actuator robot \cite{Wu2019,Lobontiu2001,Bandopadhya2010,Tang2019} and a roller made of several dielectric elastomer actuators \cite{Li2018,Firouzeh2012}.
Studies on the dynamics of many-actuator piezoelectric robots have been limited.

\begin{figure}
\centering
\subfloat[\label{fig:robot-design}]{\includegraphics[width=1\columnwidth]{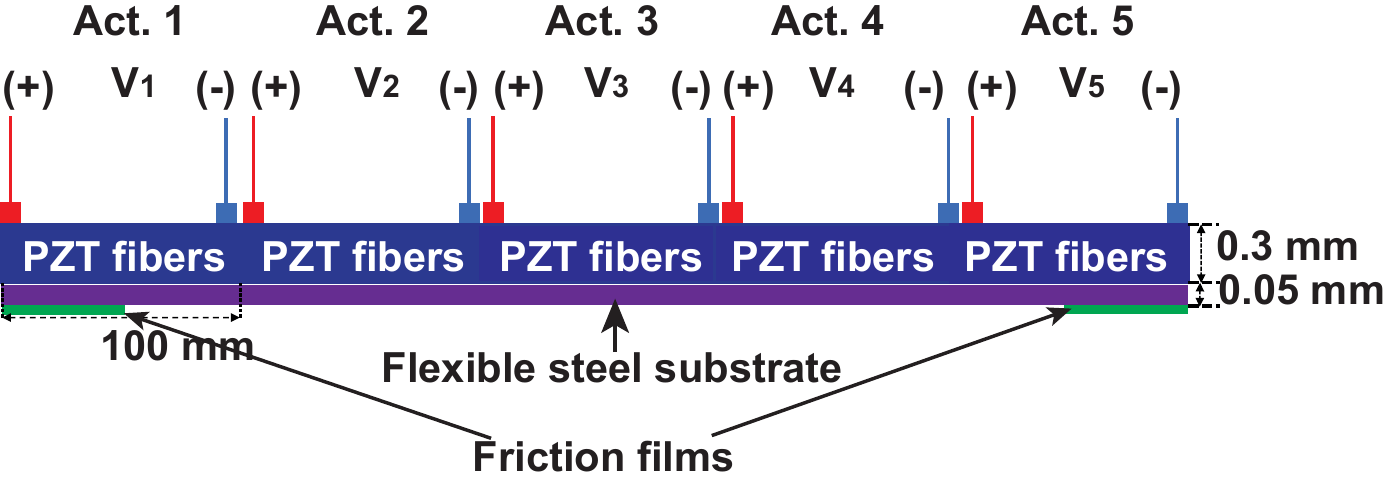}

}

\subfloat[\label{fig:actuator-design}]{\includegraphics[width=1\columnwidth]{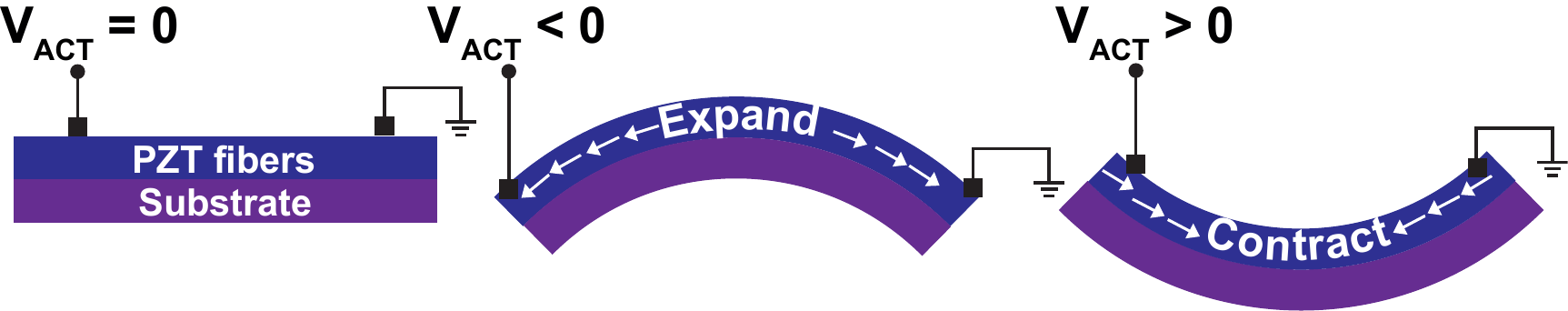}

}

\caption{(a) Cross-section of the demonstrated five-actuator soft robot prototype,
500 mm long and 20 mm wide. A high friction film of 50 mm length is applied on the underside of each end. (b) Mechanism of bending, based on piezoelectric effect, whereby an actuator unit curves concave down (up) due to expansion (contraction) under negative (positive) actuator voltage.
\label{fig:design}
}
\end{figure}

This work addresses these challenges by developing a
scalable simulation and modelling framework, generalized for a promising class of 2D soft
robots, by using a motor-link model based on a pseudo-rigid body
model, and experimentally validates the simulations.

We focus on a specific class of electrostatic soft robots which use
piezoelectric actuators. Such soft robots allow for ease of integration
and small form factors \cite{Jafferis2019} as well as fast response
times \cite{Wu2019,Ji2019}. The robot consists of a linear
array of low-cost commercially-available 100-mm-long 300-\textmu m-thick piezoelectric
composites bonded to a single 50-\textmu m-thick steel foil
(side view schematics in Fig. \ref{fig:design}(a)).
Young's modulus for the piezoelectric composites is 30 GPa, and 200 GPa of the steel foil. The demonstrated robot prototype consists of five such actuators and has a length of 500 mm and width of 20 mm. Both ends of the robot have 50-mm-long high-friction film bonded to the underside. The robot rests on a horizontal surface and is driven
by external voltage sources connected by thin compliant wires.

Fig. \ref{fig:design}(b) shows the bending mechanism of a single
actuator bonded to a steel foil. When positive (negative) voltage
is applied, the piezoelectric layer contracts (expands),
while the bottom steel, due to its stiffness, remains nearly
fixed in length. As a result, the whole structure bends concave
up (down).

\begin{figure}
\centering
\includegraphics[width=0.85\columnwidth]{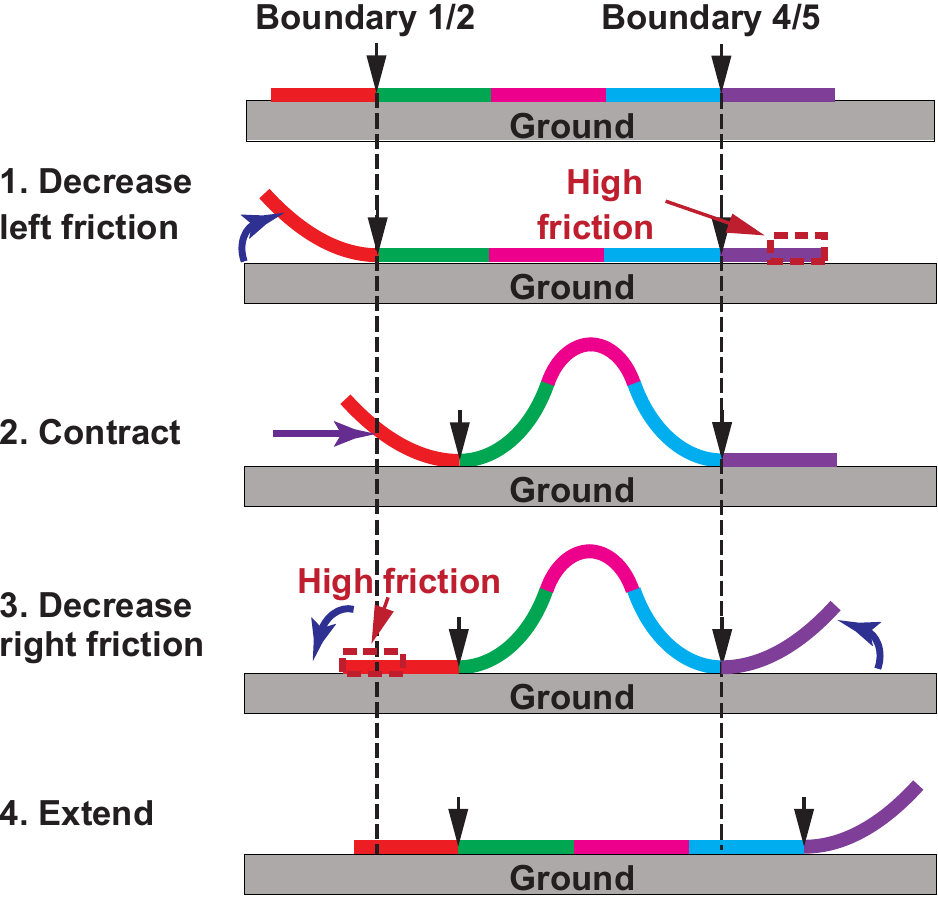}

\caption{Inchworm motion of \textquotedbl contract\textquotedbl{}
and \textquotedbl extend\textquotedbl{} cycles in 4 steps. The high-friction
films at the ends are raised and lowered on opposite ends to
create a friction asymmetry to generate motion. Each piezoelectric
actuator in the five-actuator robot is shown in by a different color.
\label{fig:inchworm-schematics}}
\end{figure}

Fig. \ref{fig:inchworm-schematics} shows a schematic view of a five-actuator robot structure
used for experimental validation in this work. Inchworm-like motion
is possible by exploiting asymmetry in friction alternating between
its two ends: in step 1, actuator \#1 (left end) is turned on to raise it to reduce friction on the left end; in step 2, actuators \#2, \#3, and \#4 are turned on; in step 3, actuator \#1 is turned off and actuator \#4 is turned on, to change the end with friction; in step 4, actuators \#2, \#3, and \#4 are turned off. The robot moves at low speed by holding a desired end fixed on the ground and then contracting/extending through its central three actuators. In addition to such motion based on robot statics, robot dynamic behaviors are also explored by operating at
higher frequencies. As described later, this enables in-place and forward
jumping motions.

\begin{figure*}
\centering
\subfloat[\label{fig:experiment:robot-experimental-setup}]{\includegraphics[width=2.0\columnwidth]{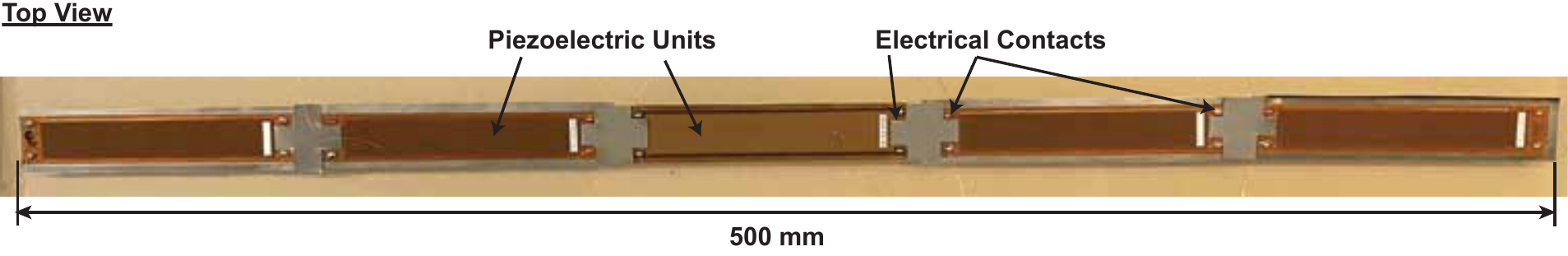}

}

\vspace{-9pt}

\subfloat[\label{fig:robot-side-view}]{\includegraphics[width=2.0\columnwidth]{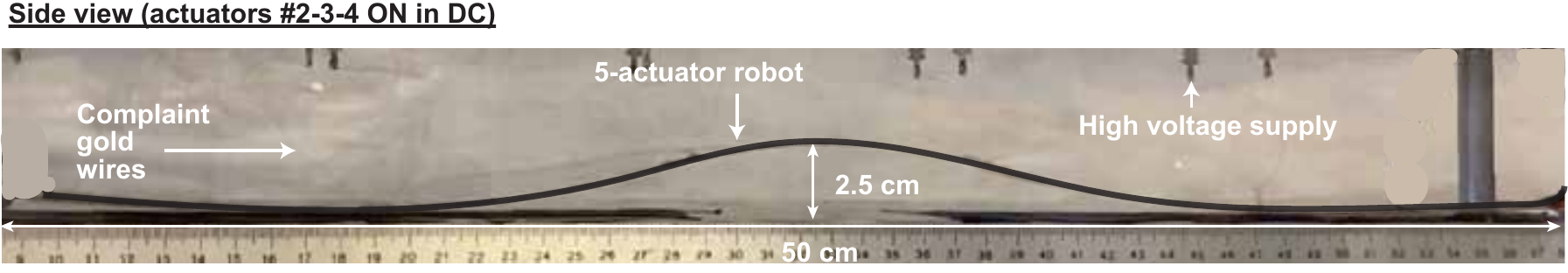}

}

\vspace{-9pt}

\subfloat[\label{fig:robot-side-view-jumping}]{\includegraphics[width=2.0\columnwidth]{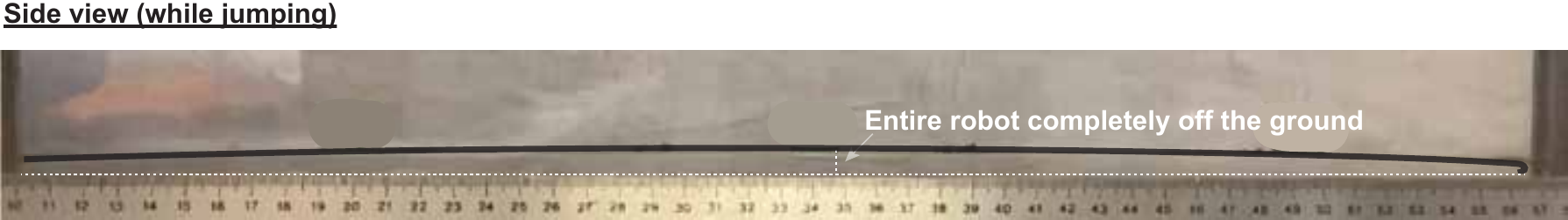}

}

\caption{Robot prototype: (a) top view; (b) side view when actuated for the
inchworm motion; (c) side view, when actuated for jumping motion.
The five-actuator robot was placed on an acrylic pad, wired to high
voltage drivers using thin and light gold wires, for real-time voltage
control.}
\end{figure*}

Fig. \ref{fig:experiment:robot-experimental-setup} shows the top
view of the robot experimental setup, Fig. \ref{fig:robot-side-view}
shows the side view with actuators \#2, \#3, and \#4  in the ON state,
and Fig. \ref{fig:robot-side-view-jumping} shows the side view, with the robot entirely off the ground in a jumping motion. Thin and light
gold wires are connected from the actuator solder pads to
high-voltage supplies, for robot control.

The paper has the following sections. Section \ref{sec:simulation-framework}
describes the PyBullet-based simulation framework and the motor-link model
of a piezoelectric actuator unit in detail. Section \ref{sec:simulator-experimental-validatio}
discusses: (1) the experimental validation of the simulation framework
for the static and dynamic analyses of a single actuator; (2)
the inchworm motion at low frequencies; and (3) symmetric in-place
jumping of the robot, a sophisticated and inherently dynamic process which cannot be captured by close-form equation models. Section \ref{sec:fast-robot-motion}
outlines the ongoing work related to the high-frequency behavior of
the robot.

\section{SIMULATION FRAMEWORK\label{sec:simulation-framework}}

\begin{figure}
\centering
\includegraphics[width=0.95\columnwidth]{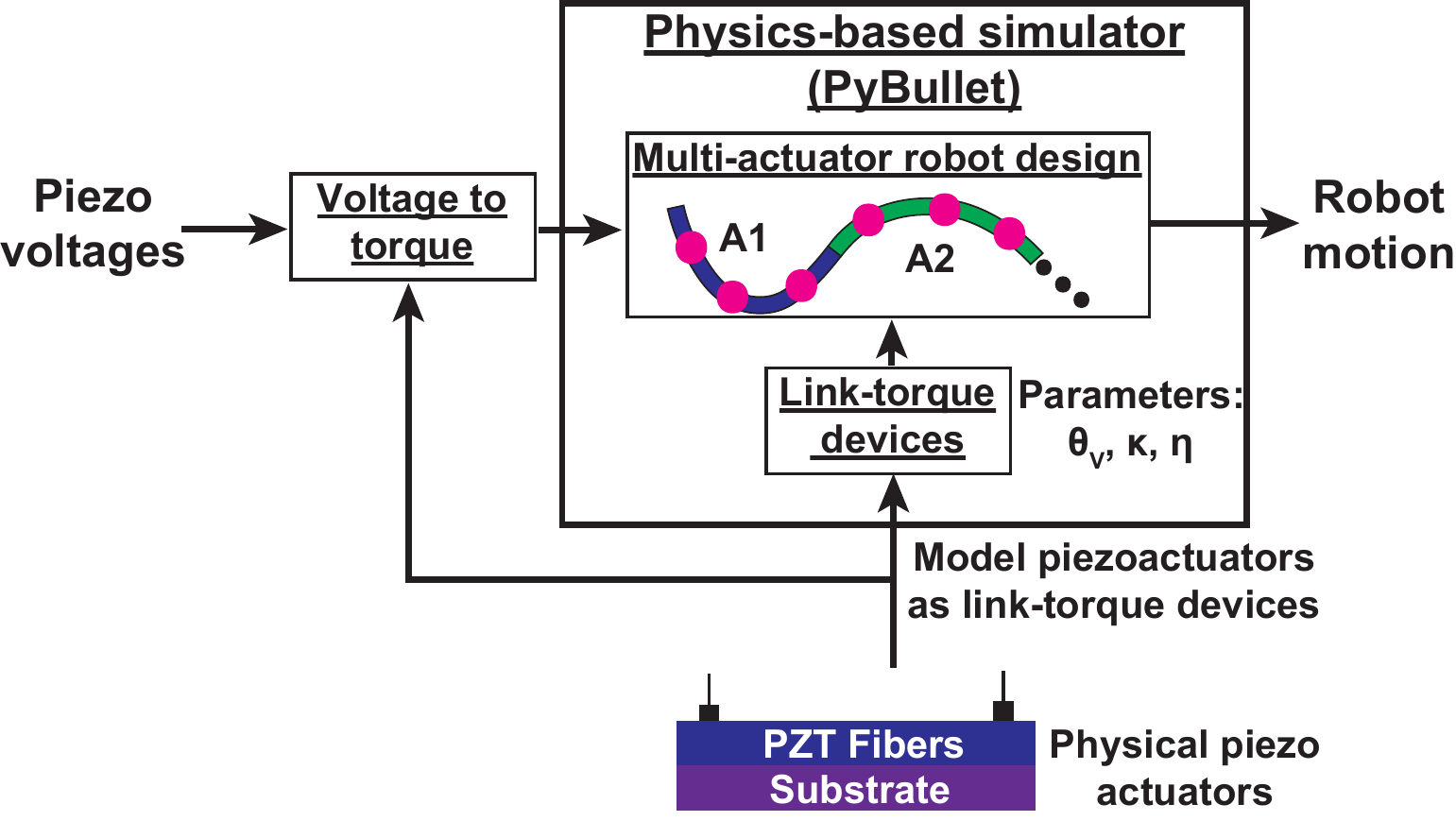}

\caption{Simulation framework block diagram, employing motor-link actuators
in PyBullet to model arrays of soft-robot actuators. \label{fig:simulation-framework-diagram}}
\end{figure}

The proposed simulation framework is integrated into PyBullet (a physics-based rigid-robot simulator \cite{coumans2019}), including
effects of gravity and friction with the ground for time-domain simulations.
Fig. \ref{fig:simulation-framework-diagram} overviews the simulation
framework. Piezoelectric actuators are modeled as link-torque devices
within PyBullet, composing a multi-actuator robot design. To use the
motor-link model, our framework converts voltages applied to the actuators
into motor torques, thereby providing robot stimuli. The resulting link-torque devices then generate forces, giving rise to the robot motions. PyBullet then solves the dynamics with the discrete Newton equations (translational and rotational) for the link-torque devices.

\begin{figure}
\centering
\includegraphics[width=0.95\columnwidth]{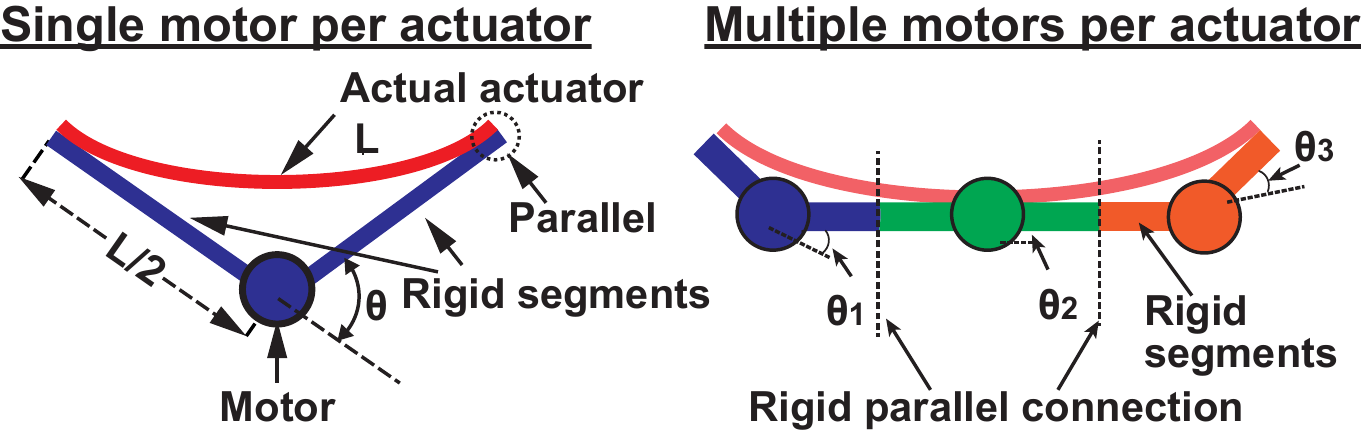}

\caption{Modelling of an actuator: an actuator is represented by a series of
motors with controlled torque connected by rigid links. Subdividing
the actuator into multiple motors with shorter rigid links improves
accuracy. \label{fig:simuulation-framework-schematics}}
\end{figure}

A key aspect of our work is the modelling of a piezoelectric actuator
unit using devices comprised of two rigid links connected at a vertex at which there is a
``motor'' (Fig. \ref{fig:simuulation-framework-schematics}). The
motor applies a torque between the links. PyBullet supports such a
motor, where the torque $\tau$ depends on the angle $\theta$ between
the two links. To represent bending stiffness, piezoelectricity, and damping, we model the torque as proportional to: (1) the deviation of the angle from the target angle of the joint; and (2) the angular velocity:

\begin{equation}
\tau=-k\left(\theta-\theta_{V}+\eta\dot{\theta}\right),
\end{equation}

\noindent where $k$ represents the ``stiffness'' of the robot; $\theta_{V}$
is a function of the voltage applied to this actuator and is determined
by piezoelectricity; and $\eta$ represents the damping. A robot with multiple actuators can be modeled in PyBullet
simply by connecting the end links of each actuator
rigidly and in parallel to that of its neighbor. Using a boundary
condition that the link ends are parallel to the piezoelectric actuator,
it can be seen that multiple motors (and shorter links), with links connected
rigidly in parallel, yield better shape modelling of an actuator than a single motor. Thus,
we modeled each actuator parametrically, with the parameter ``$m$'' corresponding to the number of motor-link units (with $m=3$ by default). Critically, for dynamic modelling, the mass of the
actuator is evenly subdivided into the links.

The simulation parameters $k$ and $\theta_{V}$ are deduced analytically,
while $\eta$ is measured experimentally in calibration experiments
employing a cantilever structure.

One can show analytically that:

\begin{equation}
k=\frac{EI}{2l}
\end{equation}

\noindent where $EI$ is the flexural rigidity of the whole robot structure, which can be determined analytically \cite{Zheng2021}, and $l$
is the link length belonging to the motor, given by

\begin{equation}
l=\frac{L}{2m},
\end{equation}

\noindent where the factor of 2 accounts for two links connected to one motor, $L$ is the length of the actuator, and $m$ is the number of
motors used to represent the actuator. Then, $\theta_{V}$ is the unloaded ``target'' angle of the motor:

\begin{equation}
\theta_{V}=\beta V/m,
\end{equation}

\noindent where $V$ is the input voltage to the actuator and $\beta$ is a constant determined by piezoelectricity, given by

\begin{equation}
\begin{aligned}
\beta &= \gamma L \\
&=\frac{d_{\text{31,1}}}{l_{\text{pitch}}}\frac{z_{1}E_{1}h_{1}}{EI}L,
\end{aligned}
\end{equation}

\noindent where $\gamma$ is the bending curvature per unit voltage \cite{Zheng2021}, $d_{31,1}$ is the piezoelectric constant of the piezoelectric
layer, $l_{\text{pitch}}$ is the distance between the neighboring
electrodes, $z_{1}$ is the position of the centerline of the piezoelectric
layer w.r.t. the neutral axis, $E_{1}$ is its Young's modulus, and $h_{1}$ is its thickness.

\section{EXPERIMENTAL VALIDATION \label{sec:simulator-experimental-validatio}}

\subsection{Single-Actuator Cantilever}

\begin{figure}
\centering
\includegraphics[width=0.95\columnwidth]{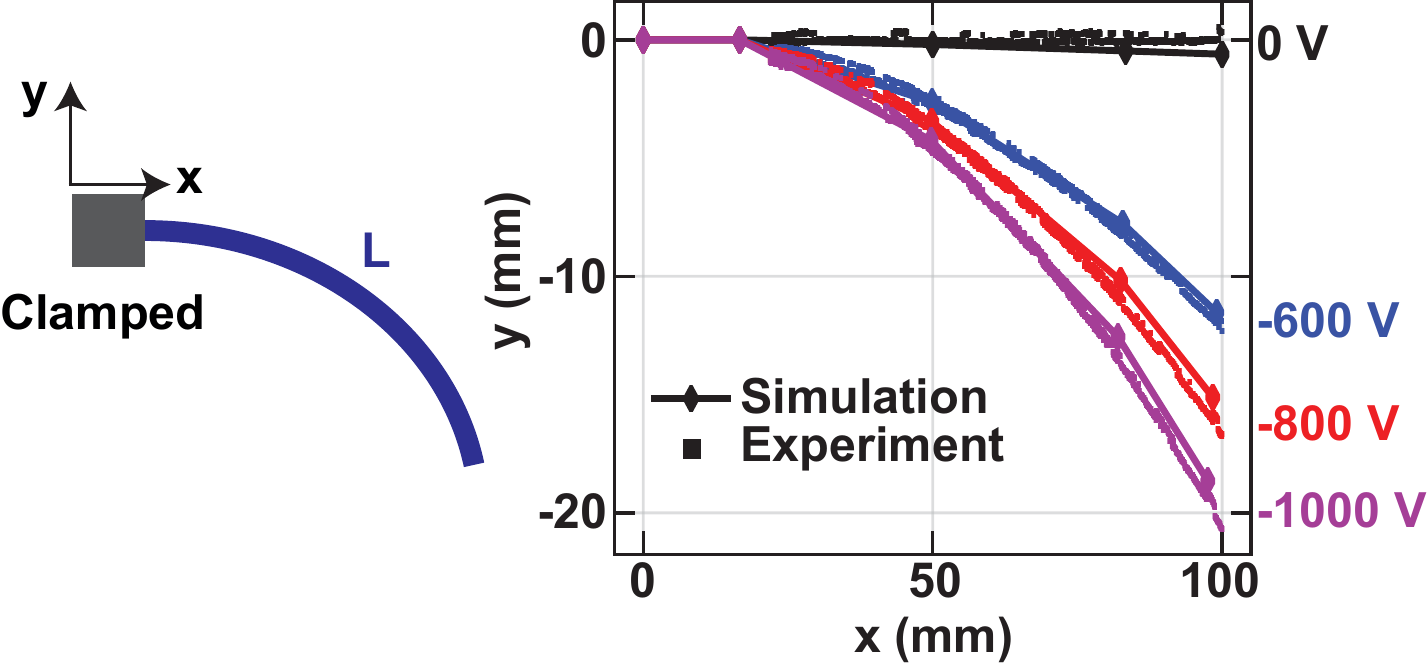}

\caption{Experimental validation of static single-actuator cantilever simulations ($m=3$) for actuator's static shapes, showing good agreement.\label{fig:cantilever-statics}}
\end{figure}

Fig. \ref{fig:cantilever-statics} shows static experimental validation
of the simulation framework for an actuator in a cantilever setting. An actuator is
clamped on its left end, and its right end is freely suspended. The actuator can bend up/down through different applied voltages.
For instance, its free end bends down by 20 mm with -1000 V applied. Good agreement
is achieved using 3-motor-link units ($m=3$). 

\begin{figure}
\centering
\includegraphics[width=0.65\columnwidth]{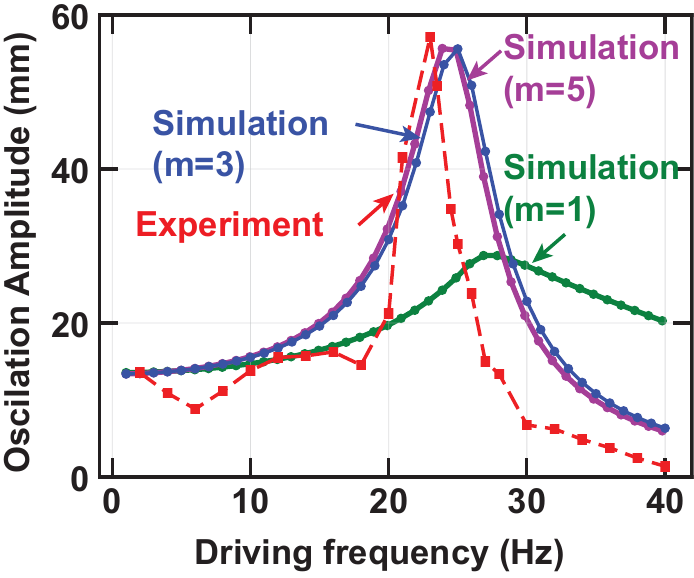}

\caption{Experimental validation of dynamic single-actuator cantilever simulation, showing achieved oscillation amplitude vs. frequency with applied sinusoidal voltages. Note that subdividing the actuator into more motor-link units improves accuracy. \label{fig:cantilever-dynamics}}
\end{figure}

By applying a step voltage vs. time, the motor damping coefficient $\eta$ was found to be 0.03
sec. Fig. \ref{fig:cantilever-dynamics} validates dynamic behaviors of
the cantilever, with an applied sinusoidal voltage between 0 V and
-1500 V. The simulated resonant frequency (25 Hz), when $m \geq 3$, is close to the
experimental 23 Hz. For the rest of the paper, $m = 3$ is used as a trade-off between precision and simulation speed.

\subsection{Robot Static Shapes and Inchworm Motion\label{subsec:inchworm-motion}}

\begin{figure}
\centering
\subfloat[\label{fig:robot-static-shape-result-1}]{\includegraphics[width=0.475\columnwidth]{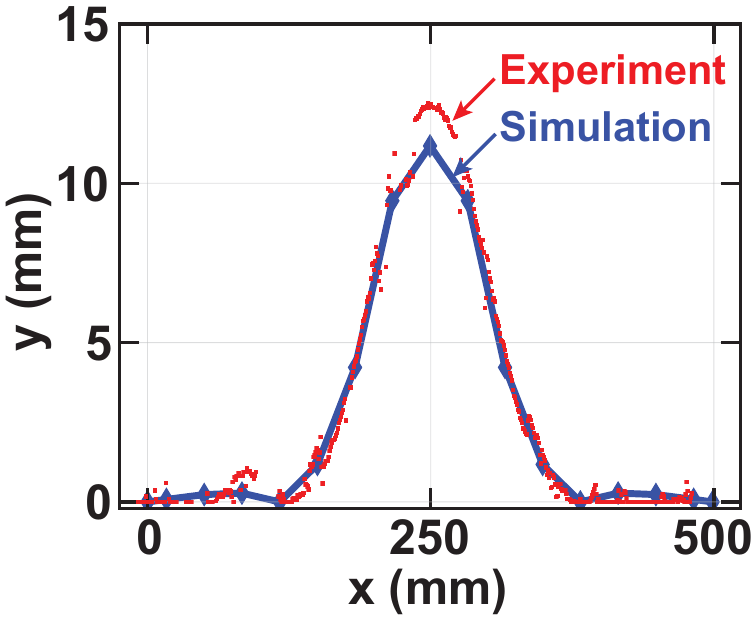}

}\hfill{}\subfloat[\label{fig:robot-static-shape-result-2}]{\includegraphics[width=0.475\columnwidth]{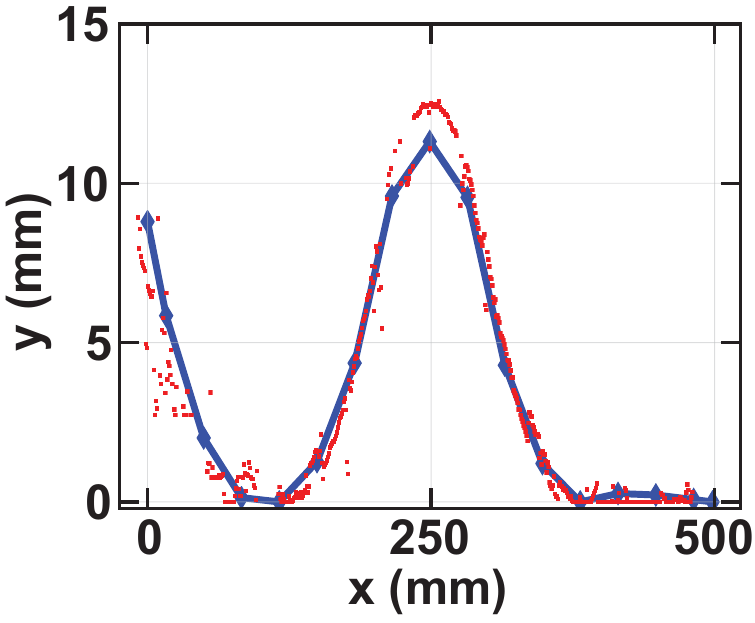}

}

\caption{Experimental validation of static shape simulations of a five-actuator
robot for different applied voltages: (a) $V_{1}=0$ V, $V_{2}=300$
V, $V_{3}=-960$ V, $V_{4}=300$ V, and $V_{5}=0$ V; (b) $V_{1}=300$
V, $V_{2}=300$ V, $V_{3}=-960$ V, $V_{4}=300$ V, and $V_{5}=0$
V. \label{fig:robot-statics}}
\end{figure}

Fig. \ref{fig:robot-statics} compares simulations and experiments
for two representative robot static shapes (chosen from the inchworm
motion steps of Fig. \ref{fig:inchworm-schematics}). Different actuators
turn on or off in each case. For experimental data, the robot shapes
are extracted from high-resolution images. Simulations and experiments
show good agreement without any curve-fitting parameters. 

\begin{figure}
\centering
\includegraphics[width=0.65\columnwidth]{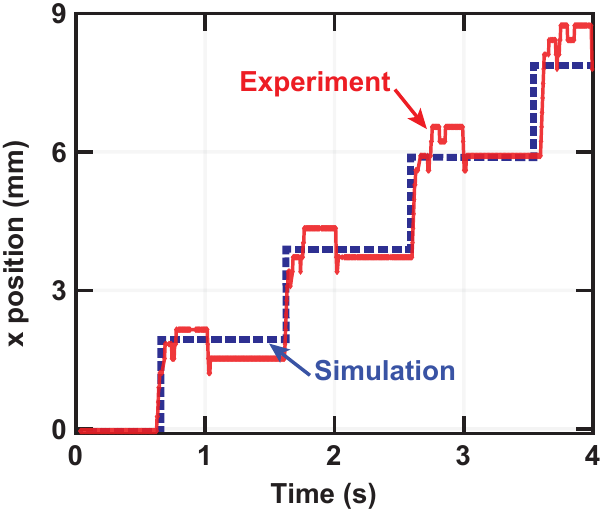}

\caption{Experimental validation of forward inchworm-motion robot simulation.
\label{fig:inchworm-result}}
\end{figure}

Fig. \ref{fig:inchworm-result} demonstrates rightward inchworm motion
of the robot (as shown in Fig. \ref{fig:inchworm-schematics}). Different
actuators turn on at different steps. The turn-on voltages are: $V_{1}=300$
V, $V_{2}=300$ V, $V_{3}=-1500$ V, $V_{4}=300$ V, and $V_{5}=300$
V. The robot moves cycle by cycle. Each cycle takes 1 s, giving overall horizontal robot
motion of 1.9 mm per cycle and average speed of 1.9 mm/s. This is again in good
agreement with the simulations.

\subsection{Robot Symmetric In-place Jumping}

\begin{figure}
\centering
\includegraphics[width=0.95\columnwidth]{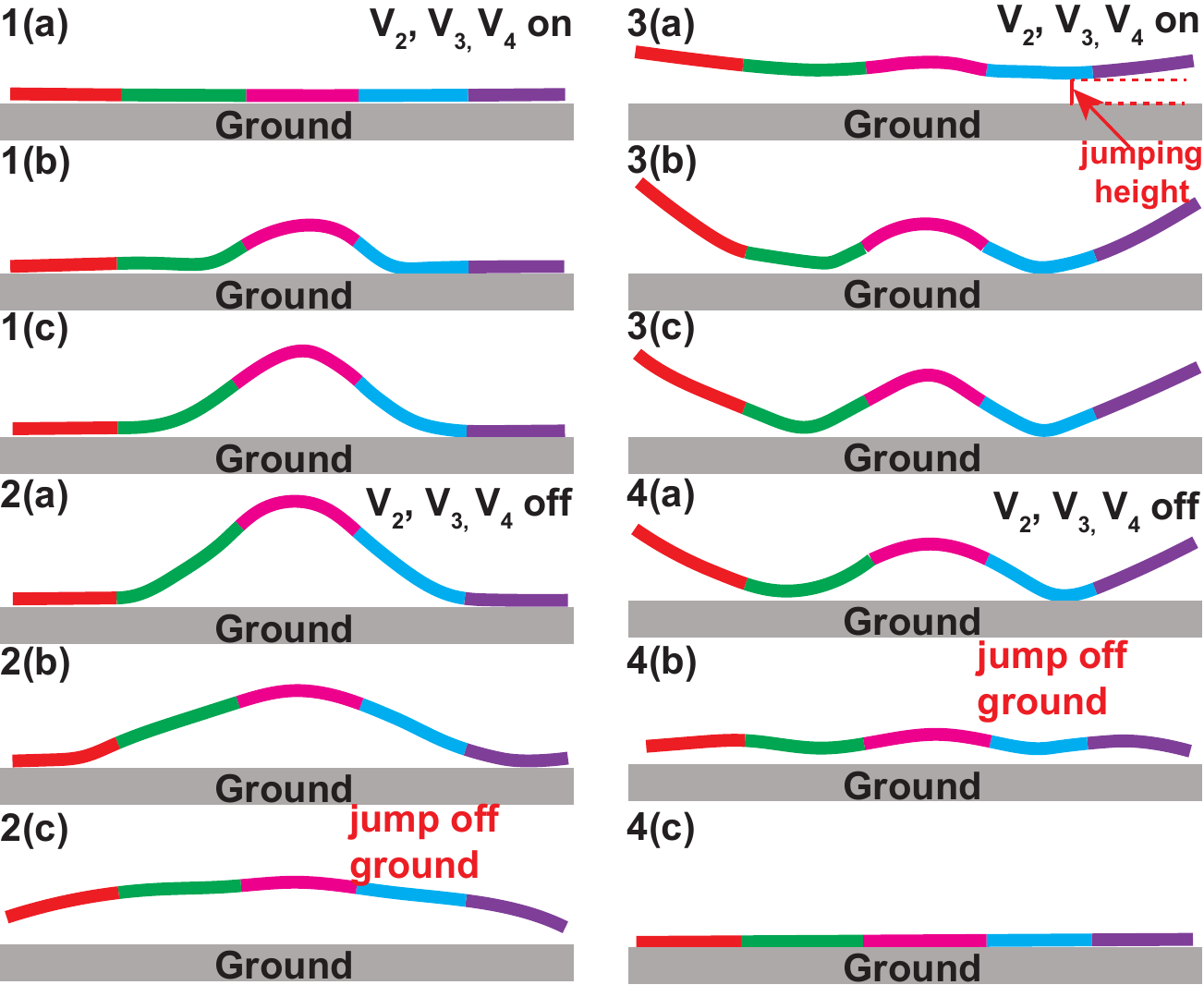}

\caption{Robot jumping observed in experiments (from high-speed cameras) due to alternating turning
$V_{2}$, $V_{3}$ and $V_{4}$ on and off. The frequency
of the voltage cycling is 14 Hz. The whole robot can jump off the ground at least 7.5 mm high, referred as ``jumping height'' (in step 3(a)). \label{fig:jumping-schematics}}
\end{figure}

When the actuator driving frequency is increased, robot dynamics play a critical role.
Here we examine a symmetric 2-phase jumping motion, where first the
central three actuators are turned on simultaneously to lift the center section,
followed by turning them off. Fig. \ref{fig:jumping-schematics}
illustrates the experimentally-observed shapes (from high-speed cameras)
in schematic form over two full periods. The following steps are observed beginning with an initially flat robot:

\begin{itemize}

\item Step 1. Actuators \#2, 3 and 4 are turned
on to lift the central section off the ground (as in Fig. \ref{fig:robot-statics}(a)).
Fig. \ref{fig:jumping-schematics} 1(a), (b), (c) show sequential experimental shapes, indicating the generation of vertical momentum.

\item Step 2. The voltages are turned off, and initially
the center of mass continues to rise. With the robot becoming
flatter, all parts are lifted off the ground.

\item Step 3. Step 1 is repeated,
but initially the robot is off the ground, so it has less time to
push itself up and generate vertical momentum compared to Step 1.

\item Step 4. $V_{2}$, $V_{3}$ and $V_{4}$ are turned off. With less initial vertical momentum, in Step 4 less height off the
ground is obtained compared to Step 2. At the end of Step 4 the robot is flat on the ground, leading back
to Step 1. 
\end{itemize}

\begin{figure}
\centering
\subfloat[\label{fig:jumping-motion-result-experiments}]{\includegraphics[width=0.475\columnwidth]{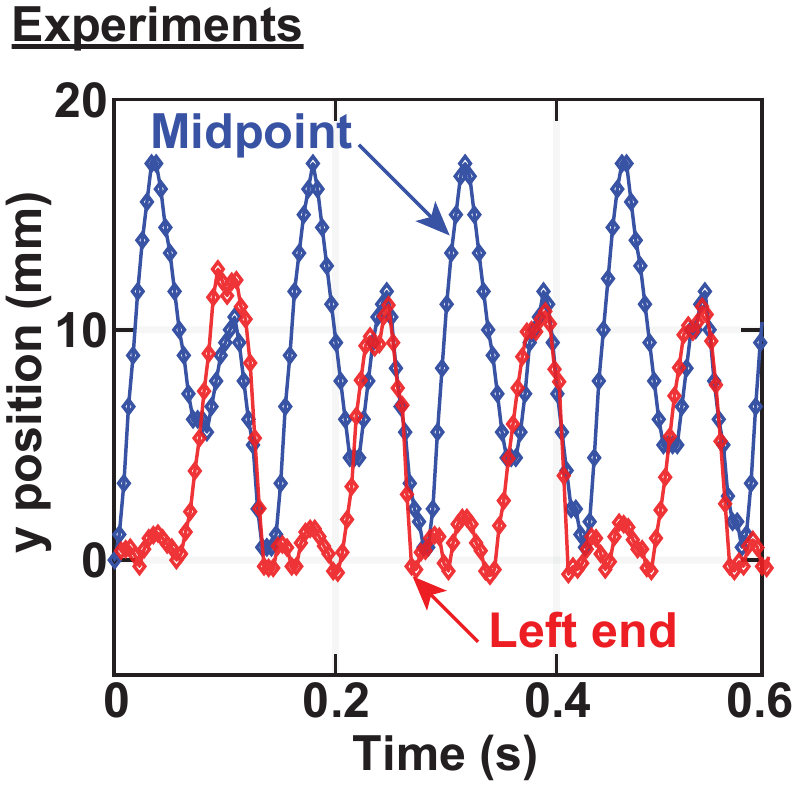}

}\hfill{}\subfloat[\label{fig:jumping-motion-result-simulations}]{\includegraphics[width=0.475\columnwidth]{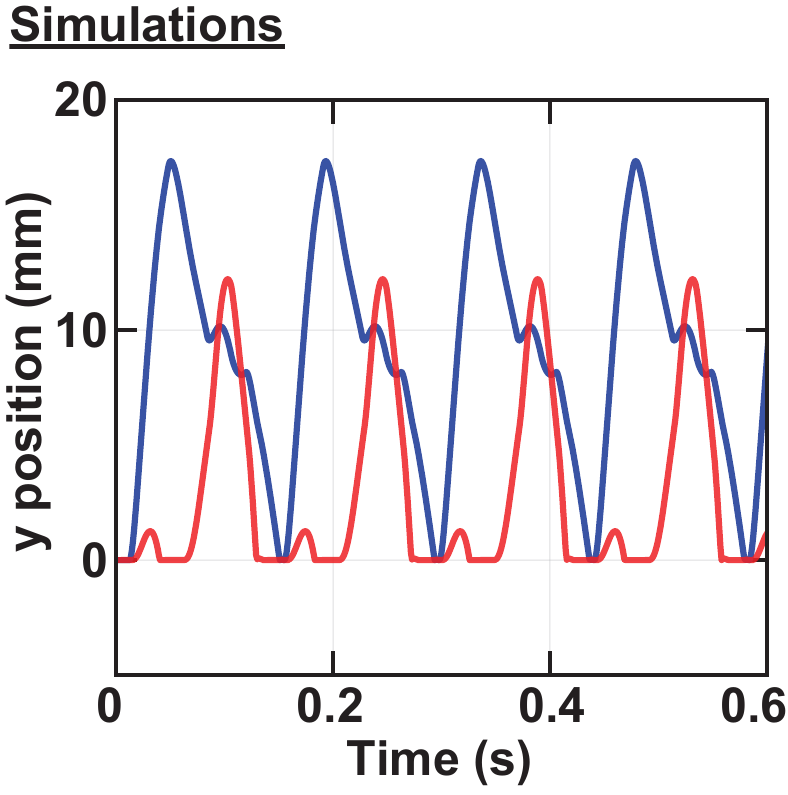}

}

\caption{Time-domain movement of the robot from (a) experiments and (b) simulations, based on tracking the vertical positions of two representative
points on the robot (blue line is robot midpoint, red line is robot
end point). The applied voltages cycle with frequency of 14 Hz, and
the frequency of periodic movement is 7 Hz. Simulations capture the
doubling of the period and the phase shift between the time of maximum
height for the middle vs. the ends.\label{fig:jumping-motion-result}}
\end{figure}

\begin{figure}
\centering
\includegraphics[width=0.75\columnwidth]{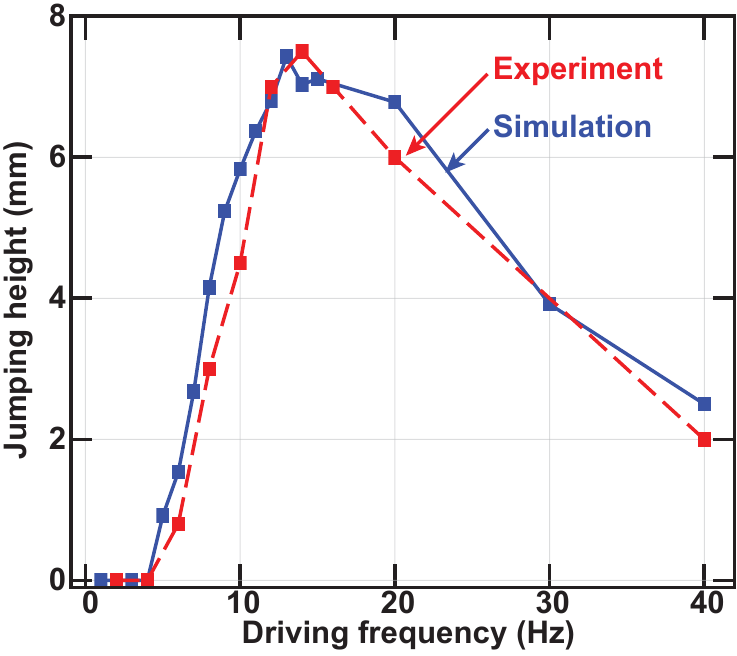}

\caption{Jumping height vs. driving frequency using three motors per actuator (m = 3) in simulations, showing excellent agreement between simulations and experiments. \label{fig:jumping-height-result}}
\end{figure}

The result is a non-linear motion, with the period of maximum height
being twice that of the applied voltage sequence (Fig. \ref{fig:jumping-motion-result}(a)).
The simulation results (Fig. \ref{fig:jumping-motion-result}(b)) accurately
capture multiple aspects of this unusual motion. Note the doubling
of the motion period compared to that of the applied voltages, and
the phase shift between the maximum height of the center of the robot
and that of its ends. Further, Fig. \ref{fig:jumping-height-result}
shows the experimental and simulated maximum height of the robot off
the ground (always measured at its lowest point) during its full cycle. The robot can jump off the ground as high as \textasciitilde 8 mm.
Remarkably good agreement between experiments and simulations is achieved without
any curve-fitting parameters.

\section{FAST MOTION EXPLORATION \label{sec:fast-robot-motion}}

\begin{figure}[h]
\subfloat[\label{fig:inchworm-frequency-results-experiments}]{\includegraphics[width=0.45\columnwidth]{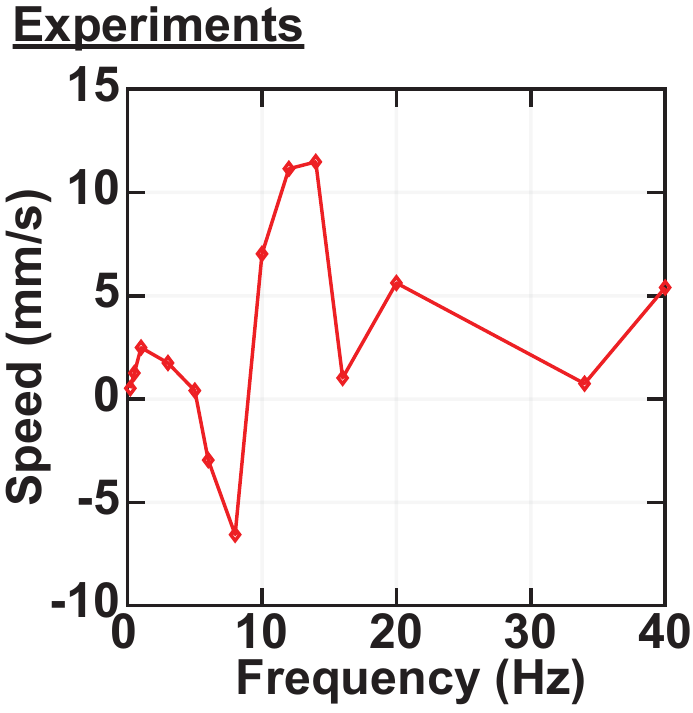}

}\hfill{}\subfloat[\label{fig:inchworm-frequency-results-simulations}]{\includegraphics[width=0.45\columnwidth]{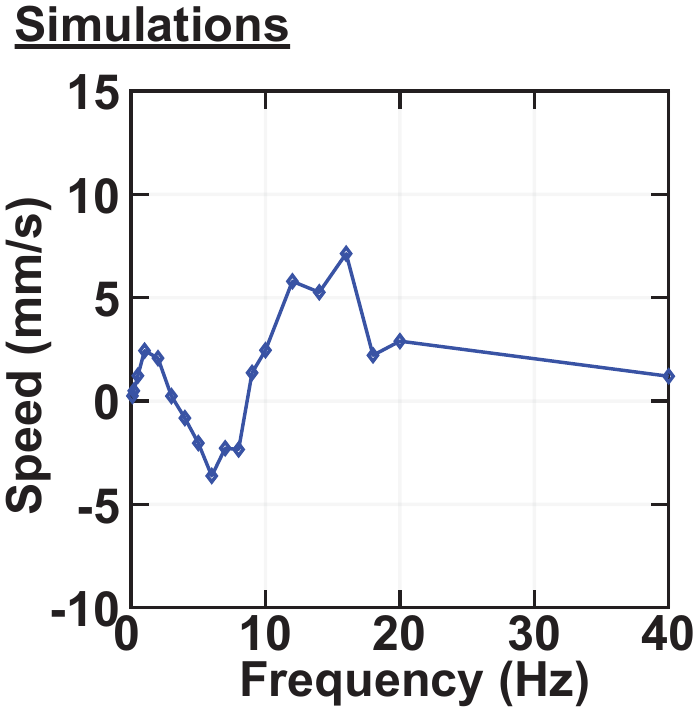}

}

\caption{Robot forward speed as a function of frequency from (a) experiments and (b) simulations. Input voltage sequences are the same as the inchworm
motion, but with a higher frequency. Simulation and
experimental results agree qualitatively, including a reversal
of direction at \textasciitilde 8 Hz, and maximum forward speed at
\textasciitilde 14 Hz. \label{fig:inchworm-frequency-results}}
\end{figure}

The symmetric applied voltages in the previous section lead to no left/right
net motion, as expected. We now further explore frequency-dependent
characteristics of the robot for the inchworm sequence of steps. The
driving frequency of the control voltages is swept from low frequencies
to high frequencies while maintaining the inchworm control-voltage sequencing
(Fig. \ref{fig:inchworm-frequency-results}). Inchworm motion (as
in Fig. \ref{fig:inchworm-result}) is observed at low frequencies
(up to 3 Hz). Beyond this, a reversal in the direction of motion is observed, maximized
at a frequency of 8 Hz for -7 mm/s. Further beyond this, at even higher frequencies,
the robot is observed to move forward again, with a peak forward speed at 14 Hz
of 12 mm/s. These frequency-dependent motions, including reversal of the movement direction, are corroborated qualitatively by the simulation, and they are currently being investigated by analyzing different vibration modes caused by different frequencies. We expect to obtain closer agreement by introducing accurately measured friction coefficients and damping factors.

\begin{figure}[h]
\centering
\includegraphics[width=1.0\columnwidth]{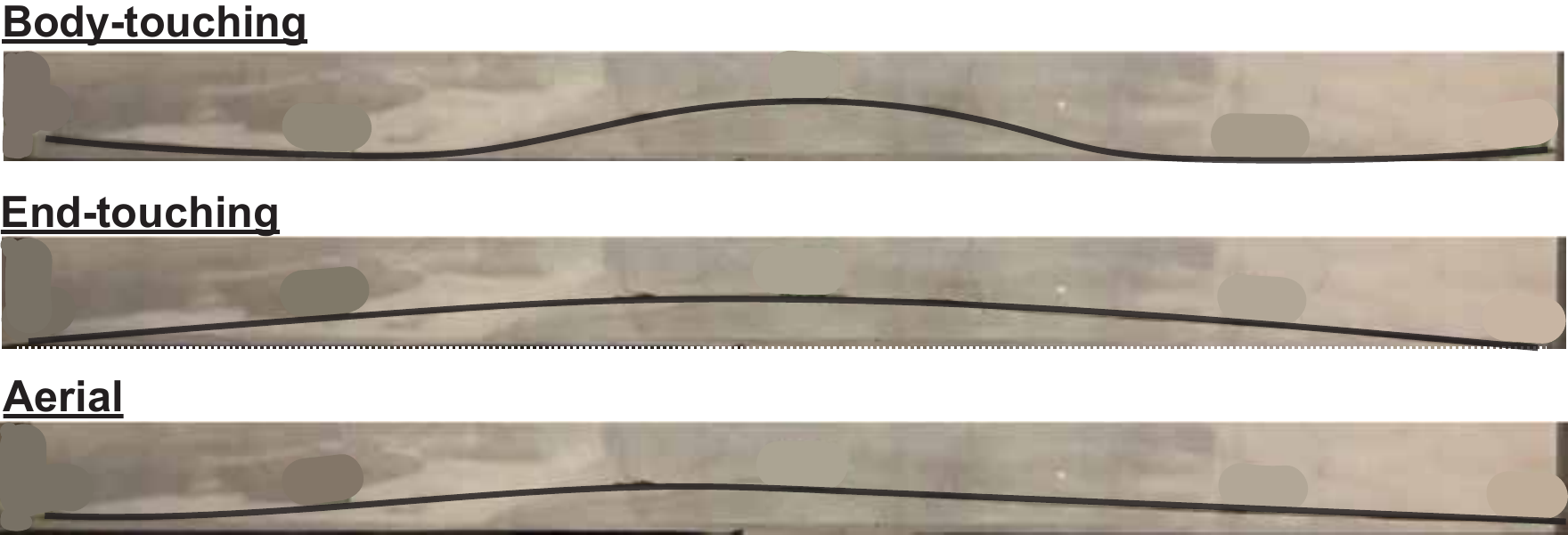}

\caption{Three representative experimental shapes for the motion at 14 Hz driving frequency.
\label{fig:key-gaits}}
\end{figure}

Fig. \ref{fig:key-gaits} shows three representative key shapes for
the motion at 14 Hz, using the voltage waveform sequence described for the inchworm motion in Fig. \ref{fig:inchworm-schematics}: body-touching, end-touching, and aerial.

We now focus on understanding how such shapes and voltage-control sequences lead to \textquotedblleft fast\textquotedblright{}
forward motion. The results of this
paper show that we can rely on simulation \textquotedblleft experiments\textquotedblright{}
to understand extremely complex non-linear behaviors of a soft robot, and move towards the design of optimal driving waveforms.

\section{CONCLUSION}

The ability of soft robots to take on rich and complex motions motivates the need for a simulation framework that captures the statics and dynamics of soft robots. This work demonstrates a PyBullet-based simulation framework that models the static and dynamic behavior of planar soft robots. A motor-torque model is used to model the soft-robot actuators, and the model parameters are based on the material properties of the actuators. The static and dynamic behavior of a single-actuator cantilever setup agree closely with the experimental results. The simulator is further validated for dynamic effects such as jumping and fast forward motion at high drive frequency where dynamic effects dominate. The model will be further used to understand more complex motion and interactions with the robot's environment.

\bibliographystyle{IEEEtran}
\bibliography{IEEEabrv, mybib}

\end{document}